\documentclass[conference]{IEEEtran}
\PassOptionsToPackage{table}{xcolor}
\usepackage{graphicx}
\usepackage{multirow}
\usepackage{amsmath}
\usepackage{amssymb,soul}
\usepackage{xcolor}
\usepackage{threeparttable}
\usepackage{booktabs}
\usepackage{colortbl}
\usepackage{bbm}
\usepackage{algorithm}
\usepackage{algpseudocode}

\definecolor{rowgray}{gray}{0.92}

\ifCLASSINFOpdf

\else

\fi

\begin{document}

% \title{\LARGE{Adaptive Additive Updates for Vision Transformers in Few-Shot Class Incremental Learning}}

\title{\LARGE{Adaptive Additive Parameter Updates of Vision Transformers for Few-Shot Continual Learning}}

\author{\normalsize Kyle Stein\(^1\), Andrew Arash Mahyari\(^{1,2}\), Guillermo Francia, III\(^3\), Eman El-Sheikh\(^3\)\\
\small \(^1\) Department of Intelligent Systems and Robotics, University of West Florida, Pensacola, FL, USA\\
\small \(^2\) Florida Institute For Human and Machine Cognition (IHMC), Pensacola, FL, USA\\
\small \(^3\) Center for Cybersecurity, University of West Florida, Pensacola, FL, USA\\
\small ks209@students.uwf.edu, amahyari@ihmc.org, gfranciaiii@uwf.edu, eelsheikh@uwf.edu

}

\maketitle

\begin{abstract}
Integrating new class information without losing previously acquired knowledge remains a central challenge in artificial intelligence, often known as catastrophic forgetting. Few-shot class incremental learning (FSCIL) addresses this by first training a model on a robust set of base classes and then incrementally adapting it in successive sessions using few labeled examples per novel class. However, this approach is prone to overfitting on limited new data, which can compromise performance and exacerbate forgetting. In this work, we propose a simple yet effective FSCIL framework that leverages a frozen Vision Transformer (ViT) backbone augmented with parameter-efficient additive updates. Our approach freezes the pre-trained ViT parameters and selectively injects trainable weights into the self-attention modules via an additive update mechanism. This design updates a small subset of parameters to accommodate new classes without sacrificing representations learned during the base session. By fine-tuning only a few parameters, our method preserves the generalizable features in the frozen ViT while reducing overfitting risk. Furthermore, as most parameters remain fixed, the model avoids overwriting knowledge when novel data batches are introduced. Extensive experiments on benchmark datasets demonstrate that our approach yields state-of-the-art performance compared to baseline FSCIL methods. Our results confirm improvements in robustness and accuracy.
\end{abstract}

% Note that keywords are not normally used for peerreview papers.

\begin{figure*}[!htbp]
\centering
\includegraphics[width=1.0\linewidth]{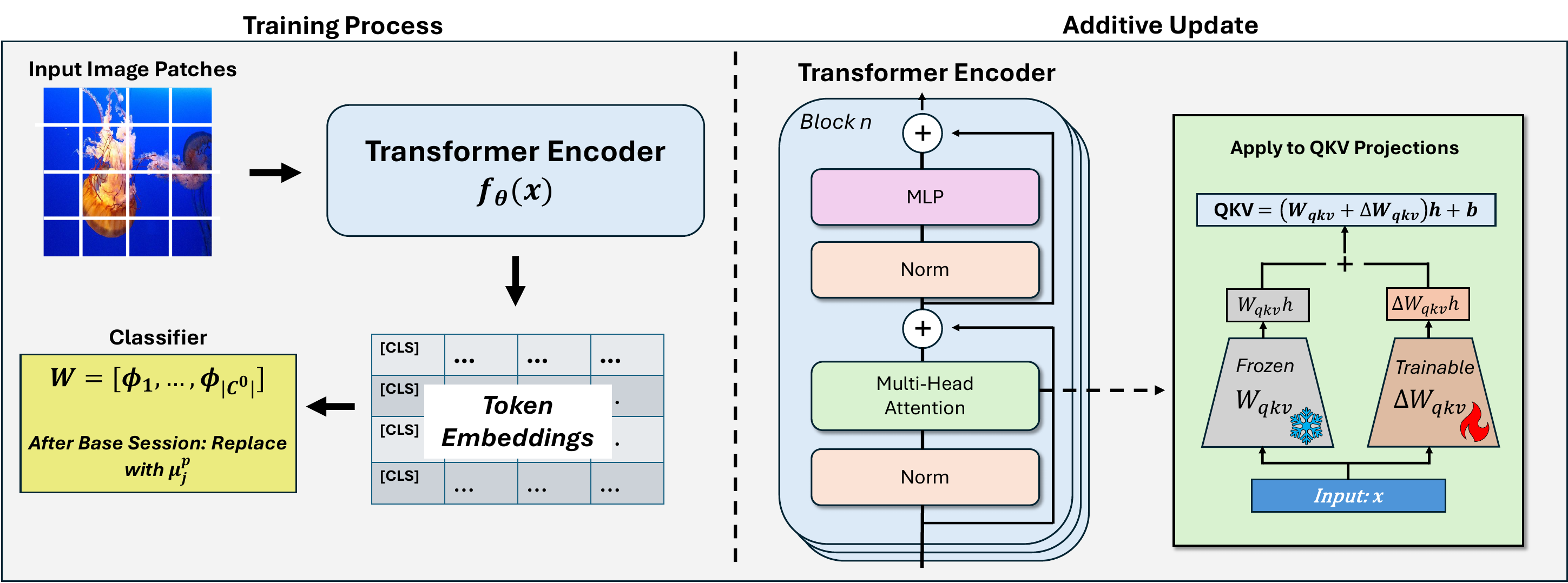}
\caption{The overall architecture of the proposed FSCIL approach.} 
\label{fig:Main_Fig}
\end{figure*}

\section{Introduction}
\label{sec:introduction}

Deep neural networks (DNNs) have shown robust performance in replicating human brain abilities, from object recognition \cite{li2022mvitv2, boutros2022elasticface} to complex language analysis \cite{devlin2018bert, radford2021learning}. However, a key assumption for their success is the availability of abundant annotated training data. In contrast, humans can rapidly recognize and retain new objects after observing them once or a few times. This discrepancy becomes more evident when models are required to incrementally learn new classes over time, a process that frequently leads to \textit{catastrophic forgetting}. Catastrophic forgetting is a paradigm that occurs when previously learned knowledge is overwritten by newly acquired information \cite{french1999catastrophic}. Class-incremental learning (CIL) \cite{zhou2024class, zhou2024revisiting} aims to address this challenge by enabling models to learn new classes over time while still retaining knowledge from previously encountered classes. 

Although CIL has shown promise in mitigating catastrophic forgetting, existing methods still require a substantial number of samples for each new incremental class \cite{meng2024diffclass, liu2025class}. Few-shot class-incremental learning (FSCIL) seeks to overcome this limitation by enabling models to incrementally learn new classes from only \textit{a few labeled samples}, while still preserving the knowledge of previously encountered classes \cite{tao2020few, tian2024survey}. FSCIL consists of a base session and subsequent incremental sessions. During the base session, the model is trained on large amounts of data, enabling it to learn fine-grained features and establish robust prior knowledge that can be effectively transferred to new classes. In contrast, incremental sessions introduce new classes with only a few samples each, which may cause the model to overfit, since these limited examples might not accurately represent the true distribution of the new classes.

Earlier works \cite{kim2023warping, yang2023neural, oh2024closer, zou2024compositional} have used various ResNet models \cite{he2016deep} to mitigate overfitting; however, these models often struggle to transfer knowledge from the base session to incremental sessions. Recent studies \cite{park2024pre, yuan2025prompt, liang2024inflora} have adopted the more powerful Vision Transformers (ViT) \cite{dosovitskiy2020image} to enhance base session performance. Nonetheless, since ViTs are pre-trained on large-scale datasets like ImageNet \cite{deng2009imagenet}, they are particularly susceptible to catastrophic forgetting when fine-tuned on only a few novel examples. In few-shot and incremental learning scenarios, the limited and distributionally different fine-tuning data can easily lead to overwriting the robust pre-trained features, thereby amplifying this risk of catastrophic forgetting.

To prevent performance degradation in incremental sessions, a promising solution lies in parameter-efficient fine-tuning (PEFT). PEFT is a strategy for adapting a large, pre-trained model to a new task by making only minimal changes to its parameters \cite{ding2023parameter}. A prominent submethod, prompt tuning \cite{li2021prefix, liu2024few} \cite{stein2025visual}, adapts the model by prepending context-specific learnable tokens into the model's input layer. Previous FSCIL methods have employed prompt tuning techniques \cite{liu2024few}; however, prompt tuning has been shown to converge slower and more difficult to optimize than other PEFT methods \cite{ding2023parameter, zaken2021bitfit}. In contrast, alternative PEFT methods \cite{hu2022lora} operate on the internal layers of the backbone model, such as the self-attention modules within transformer blocks, by injecting learnable matrices while keeping the pre-trained weights frozen. This approach avoids the overhead of processing additional input tokens and mitigates the optimization issues commonly encountered with prompt tuning.

Motivated by these insights, we propose a novel parameter-efficient approach to address catastrophic forgetting and overfitting in FSCIL. We leverage a pre-trained VIT with a \textbf{key contribution}: an additive update mechanism. Unlike extensive fine-tuning or traditional prompt tuning, which require re-training large portions of the model or adding extra modules that are difficult to optimize, our method selectively updates only the self-attention layers within the frozen ViT backbone. This targeted adaptation preserves the robust pre-trained representations while enhancing computational efficiency and ease of integration. Our approach shares the same additive parameters across all blocks of the transformer, which reduces the number of additive parameters significantly while preserving the performance. In addition, we refine the classifier by recalibrating each class-specific weight vector to better align with the relevant feature distribution. Together, these components enable our approach to effectively mitigate catastrophic forgetting and overfitting in FSCIL, achieving state-of-the-art performance over baseline methods.

\section{Related Work}
\label{sec:related-work}

\subsection{Few-Shot Class-Incremental Learning}
Classic CIL \cite{zhou2024class, liu2025class} continually learns new classes using ample data per incremental session, while FSCIL \cite{tao2020few} mimics human learning by adapting to new classes with only a few samples after extensive base training.  One common perspective divides methods into metric-based and fine-tuning based strategies. The metric-based methods \cite{yang2023neural, chi2022metafscil, zhou2022few} focus on robust prototype representation and similarity metrics, similar to traditional few-shot learning techniques \cite{snell2017prototypical, vinyals2016matching, stein2025transductiveoneshotlearningmeet}. Fine-tuning based approaches \cite{peng2022few, hersche2022constrained} leverage extensive base class training to initialize the model, followed by careful fine-tuning on novel classes. Recently, the focus has shifted from shallower models such as ResNet for feature extraction to leveraging the more powerful ViT \cite{park2024pre, liu2024few, li2025continuous}. However, fine-tuning powerful models like ViT for incremental sessions is challenging because of catastrophic forgetting. Existing methods mitigate catastrophic forgetting by freezing the pre-trained ViT backbone to preserve its learned representations and incorporating prompt tuning-based approaches \cite{park2024pre, liu2024few}. However, prompt tuning is known to converge slower and is more difficult to optimize \cite{ding2023parameter, zaken2021bitfit}.  In contrast, our approach shares updates to the self-attention layers within transformer blocks of the frozen backbone, effectively preventing catastrophic forgetting while keeping the model adaptation strategy simple.

\subsection{Parameter-Efficient Fine-Tuning}
The goal of PEFT is to adjust a limited number of large model parameters while keeping the remainder of the parameters unchanged \cite{han2024parameter}. PEFT can be summarized into reparameterization-based methods, specification-based methods, and addition-based methods \cite{ding2023parameter}. Reparameterization-based methods operate on insight that the adaptations required for many tasks reside in a low-dimensional subspace of the full parameter space \cite{hu2022lora, zhang2023adalora}. Specification-based methods do not add additional parameters to the model and optimize a pre-selected subset of the model’s existing parameters, such as updating strictly bias terms in a pre-trained network \cite{zaken2021bitfit}. Addition-based methods introduce extra parameters into the network without modifying the original pre-trained weights \cite{stein2025visual, gao2020making}. In our work, we propose a targeted addition-based approach to the self-attention layers of transformer blocks. This approach adapts the model by adding small adjustments only in the self-attention layers, leaving the main pre-trained parameters intact. Moreover, we share a single additive parameter across all selected transformer blocks, further reducing the number of additional parameters required for adaptation.

\vspace{-2mm}
\section{Preliminaries}
\label{sec:preliminaries}
In FSCIL, a model is trained sequentially over multiple sessions, denoted as
${D} = \{ D^{(0)}, D^{(1)}, \dots, D^{(T)} \}$, where the base session is denoted as $D^{(0)}$ and $T$ is the total number of incremental tasks. Each incremental session $t$ provides a training dataset: $D^{(t)} = \{ (x_i, y_i) \}_{i=1}^{|D^{(t)}|}$, where $x_i$ is an input sample and $y_i$  is its corresponding label. In the base session $D^{(0)}$, a comprehensive label space $C^{(0)}$ is provided along with ample data for each class, thereby establishing a robust foundation for base model training. However, for each incremental session $D^{(t)}$, with $1 \le t \le T$, only a few training samples per class are available, arranged in an $N$-way $K$-shot format, where $N$ new classes are introduced with $K$ examples each. Notably, classes introduced in different sessions are entirely distinct: for any $t \neq t'$, the sets $C^{(t)}$ and $C^{(t')}$ are disjoint. During inference, the model is evaluated on all classes encountered up to that point. Specifically, after training session $t$, testing is performed on a dataset  $D^{(\leq t)}_{\text{test}}$, which consists solely of examples whose labels belong to the cumulative set  $C^{(\leq t)} = \bigcup_{j=0}^{t} C^{(j)}$. The model's goal is to incrementally incorporate new classes from only a few examples, while maintaining its ability to accurately classify all previously learned classes.

\section{Proposed Method}
\label{sec:proposed-method}
To address the limitations of existing approaches for FSCIL, we propose a novel framework that adapts a pre-trained ViT backbone to new classes by fine-tuning few parameters in base session, \textit{i.e.,} PEFT. Our method leverages additive updates to the self-attention layers within each transformer block, while keeping pre-trained parameters frozen. Specifically, we apply updates to the query, key, and value (QKV) projection matrices. These projection matrices capture the contextual relationships among patches in each input image through positional embeddings and self-attention. This design enables the model to effectively incorporate new class information from only a few labeled examples without compromising the robust feature representations learned during the base session. In addition, we refine the classifier by recalibrating each class-specific weight vector to better align with the evolving feature distribution. The overall architecture of the proposed design is shown in Figure~\ref{fig:Main_Fig}.

%To address the limitations of existing approaches for FSCIL, we propose a novel framework that adapts a pre-trained ViT backbone by fine-tuning a small set of parameters (i.e., PEFT) during the base session, enabling effective adaptation to new classes. Our method leverages additive updates to the self-attention layers within each transformer block while keeping the pre-trained weights frozen. Specifically, we modify the query, key, and value (QKV) projection matrices. This design enables the model to capture robust contextual relationships among image patches through positional embeddings and self-attention, allowing it to incorporate new class information from only a few labeled examples without compromising the learned feature representations. Furthermore, in subsequent incremental sessions, the backbone remains fixed while the classifier is refined by recalibrating its class-specific weight vectors to better align with the evolving feature distribution. The overall architecture of the proposed design is shown in Figure~\ref{fig:Main_Fig}.

\subsection{Feature Extraction}
Following prior works \cite{park2024pre, liu2024few, li2025continuous}, our classifier employs a pre-trained ViT model \(f_\theta(\cdot)\) as the feature extractor. The ViT is initialized with weights pre-trained on large-scale datasets (i.e., ImageNet-21K), and its original classification head is removed to yield raw feature representations. Specifically, given an input image \(x \in \mathbb{R}^{C \times H \times W}\), the image is first divided into patches and embedded into a sequence. A classification token ([CLS]) is prepended to this sequence, and positional encodings are added. The transformer encoder then processes this sequence to output $h = f_\theta(x) \in \mathbb{R}^{N \times D}$, where $h$ denotes the sequence of token embeddings, $N$ is the total number of tokens, and $D$ is the embedding dimension. The [CLS] token, which captures the global image representation, is extracted and denoted as $z = h_{\text{cls}} \in \mathbb{R}^{D},$ and is subsequently normalized to obtain the final representation from the transformer encoder.

\subsection{Additive Update Mechanism in Self-Attention Layers}

To facilitate controlled adaptation during the base session, we introduce an additive update mechanism that specifically targets the QKV projections within the self-attention modules of each transformer block. In a standard ViT, these projections are computed as follows:
\begin{equation}
\begin{aligned}
Q &= W_q\, h + b_q, \\
K &= W_k\, h + b_k, \\
V &= W_v\, h + b_v.
\end{aligned}
\end{equation}

Here, \(W_q\), \(W_k\), and \(W_v\) are the pre-trained weight matrices, and \(b_q\), \(b_k\), and \(b_v\) are the corresponding bias terms. In our approach, each projection is modified by adding a trainable update to the corresponding weight matrix while retaining the original trainable bias:
\begin{equation}
\begin{aligned}
Q &= (W_q +  \Delta W_q)\, h + b_q, \\
K &= (W_k +  \Delta W_k)\, h + b_k, \\
V &= (W_v +  \Delta W_v)\, h + b_v.
\end{aligned}
\end{equation}

Specifically, $\Delta W_q$, $\Delta W_k$, and $\Delta W_v$ are the trainable additive parameters (initialized to zero).  Notably, we \textbf{share a single $\Delta W_q$, $\Delta W_k$, and $\Delta W_v$ across all selected transformer blocks}, which significantly reduces the number of parameters being updated throughout training process \cite{takase2021lessons}. %\ar{a question, I understand $\Delta W$ is the same across all blocks, but isn't it different for q, k, and v? so shouldn't we have $\Delta W_q$, $\Delta W_k$ and $\Delta W_v$?   }

%During the base session, both $\Delta W$ and the bias terms are updated via gradient descent, while the original weight matrices remain frozen. Due to the same $\Delta W$ being shared across all adapted transformer blocks, the gradients from each block are aggregated to update this single parameter. In incremental sessions, however, all parameters of the feature extractor are frozen to preserve the representations learned during the base session. 

\subsection{Training Procedure} 
\noindent \textbf{Base Session.} During the base session, the network is trained on a set of base classes using a softmax cross-entropy loss computed on logits derived from cosine similarity. For each training sample $x_i$, the feature extractor $f_\theta(\cdot)$ produces a final feature vector $z = h_{\text{cls}} \in \mathbb{R}^{D},$ from the [CLS] token which is then normalized. The classifier is implemented as a linear layer with a weight matrix:
%\ar{you introduced h as features in A}

\vspace{-1mm}
\begin{equation}\label{eq:classifier}
\begin{aligned}
W = [\phi_1, \phi_2, \ldots, \phi_{|\mathcal{C}^{(0)}|}],
\end{aligned}
\end{equation}

\noindent where each $\phi_j \in \mathbb{R}^D$ is a learned weight vector of class $j$. The classifier maps an input's normalized feature $z_i$ to a set of logits by computing the cosine similarity between $z_i$ and each normalized classifier weight $\hat{\phi}_j$ (where $\hat{\phi}_j = \phi_j / ||\phi_j||_2$). During base training, these classifier weights are learned via gradient descent to align the feature embeddings with the correct class. The cosine similarity is defined as:

\vspace{-1mm}
\begin{equation}
\text{sim}(z_i, \phi_j) = z_i^\top \hat{\phi}_j.
\end{equation}

To obtain the logit for class $j$, we scale this similarity by a temperature parameter $\tau$:

\vspace{-1mm}
\begin{equation}
\ell_{i,j} = \tau\, \text{sim}(z_i, \phi_j).
\end{equation}

In the softmax cross-entropy loss, $\exp(\ell_{i,y_i})$ corresponds to the exponential of the logit for the ground-truth class $y_i$ of sample $i$. The loss over a batch of size $B$ is then defined as:

\vspace{-3mm}
\begin{equation}
\begin{aligned}
\mathcal{L}_{\text{ce}} = \frac{1}{B} \sum_{i=1}^{B} -\log \left(\frac{\exp\bigl(\ell_{i,y_i}\bigr)}{\sum_{j=1}^{|\mathcal{C}^{(0)}|} \exp\bigl(\ell_{i,j}\bigr)}\right).
\end{aligned}
\end{equation}

\noindent Minimizing ${L}_{\text{ce}}$ encourages each feature vector $z_i$ to align with its corresponding classifier weight while distancing it from those of other classes.

In addition to updating the classifier weights, the shared additive parameters \(W_q\), \(W_k\), and \(W_v\), as well as the associated bias terms are also updated. During base training, the gradients from the cross entropy loss are backpropagated through all adapted blocks. Furthermore, since we share the same additive updates in each block, gradients are aggregated to update this single parameter via gradient decent. Meanwhile, the original pre-trained weight matrices $W_q$, $W_k$, and $W_v$ remain frozen. In incremental sessions, all parameters of the feature extractor, including the additive updates, are frozen to preserve the representations learned during the base session. Only the classifier is updated for new classes.

\vspace{2mm}
\noindent \textbf{Classifier Replacement.} 
After the base session training is completed, we refine the classifier via \emph{classifier replacement} \cite{oh2024closer}. Recall that the classifier is implemented as a linear layer with a weight matrix from Eq.~\ref{eq:classifier}, where each $\phi_j \in \mathbb{R}^D$ is the learned weight vector for class $j$. Previous works \cite{oh2024closer, park2024pre, liu2024few} have shown that reinitializing these weight vectors to better capture the underlying feature distributions can avoid overfitting and catastrophic forgetting. To reinitialize these weight vectors for each base class $j$, we define a learnable parameter $\mu_j$. Let  $\mathcal{S}_j = \{ i \mid y_i = j \}$ denote the set of indices for image samples belonging to class $j$. The parameter $\mu_j$ for each class $j$ is learned by solving the following optimization problem: 

\vspace{-3mm}
\begin{equation} \label{eq:prototype}
\begin{aligned}
\min_{\mu_j} {1 \over 2} \sum_{i \in \mathcal{S}_j}{\|\mu_j-z_i\|^2}.
\end{aligned}
\end{equation}

Once the optimization is solved, we normalize $\mu_j$ and update the corresponding classifier weight $\phi_j$ with the normalized solution. This step ensures that the classifier's weights accurately represent the aggregate features of their respective classes.

\begin{table*}[t!]
\centering
\caption{The performance of sessions on CUB200.}
\label{tab:cub200-sessions}
\begin{threeparttable}
\renewcommand{\arraystretch}{1.15}
\setlength{\tabcolsep}{5pt}
\begin{tabular}{lccccccccccccc}
\toprule
\rowcolor{rowgray}
\textbf{Method} & 
\textbf{$S_{\mathrm{base}}$} & 
\textbf{1} & 
\textbf{2} & 
\textbf{3} & 
\textbf{4} & 
\textbf{5} & 
\textbf{6} & 
\textbf{7} & 
\textbf{8} & 
\textbf{9} & 
\textbf{$S_{\mathrm{last}}$} & 
\textbf{$S_{\mathrm{avg}}$} &
\textbf{PD} \\
\midrule

CEC\(\ddagger\) \cite{zhang2021few}
& 75.40 & 73.23 & 72.00 & 68.70 & 69.35 & 67.78 & 67.01 & 66.40 & 65.78 & 65.57 & 65.70 & 72.41 & 9.70 \\

L2P\(\ddagger\) \cite{wang2022learning}
& 44.97 & 30.28 & 27.21 & 24.44 & 22.41 & 20.81 & 19.47 & 18.19 & 17.16 & 16.26 & 15.41 & 24.99 & 29.56 \\

DualPrompt\(\ddagger\) \cite{wang2022dualprompt}
& 53.37 & 45.99 & 41.15 & 37.33 & 34.32 & 31.57 & 29.44 & 27.58 & 25.92 & 24.55 & 23.25 & 36.30 & 30.12 \\

NC-FSCIL\(\ddagger\) \cite{yang2023neural}
& 78.49 & 71.52 & 65.54 & 60.30 & 55.81 & 51.96 & 48.72 & 45.78 & 43.18 & 40.92 & 38.80 & 57.92 & 39.69 \\

WaRP\(\ddagger\) \cite{kim2023warping}
& 67.74 & 64.21 & 61.06 & 57.80 & 55.78 & 53.81 & 52.82 & 51.61 & 50.13 & 50.02 & 49.36 & 55.85 & 18.38 \\

PL-FSCIL\(\dagger\)\cite{yuan2025prompt}
& 85.16 & 85.40 & 82.75 & 75.22 & 77.22 & 73.25 & 72.39 & 70.24 & 67.97 & 68.33 & 69.86 & 75.25 & 15.30\\

PriViLege\(\ddagger\) \cite{park2024pre}
& 82.21 & 81.25 & 80.45 & 77.76 & 77.78 & 75.95 & 75.69 & 76.00 & 75.19 & 75.19 & 75.08 & 77.50 & 4.71\\

\midrule

\rowcolor{rowgray}
\textbf{Proposed Method\(\dagger\)}
& 87.47 & 85.61 & 84.46 & 82.70 & 82.70 & 81.14 & 80.60 & 81.02 & 80.47 & 80.67 & 80.86 & 82.52 & 6.61 \\

\midrule

\rowcolor{rowgray}
\textbf{Proposed Method\(\ddagger\)}
& \textbf{88.41} & \textbf{87.02} & \textbf{86.27} & \textbf{84.90} & \textbf{85.06} & \textbf{83.45} & \textbf{83.46} & \textbf{83.98} & \textbf{83.81} & \textbf{84.02} & \textbf{84.26} & \textbf{84.97} & \textbf{4.15} \\

\bottomrule
\end{tabular}
\begin{tablenotes}
\footnotesize
\item \textbf{Note:} The columns represent the base session (\(S_{\mathrm{base}}\)), incremental sessions (1--9), the final session (\(S_{\mathrm{last}}\)), the average accuracy (\(S_{\mathrm{avg}}\)), and the performance drop (PD) between \(S_{\mathrm{base}}\) and \(S_{\mathrm{last}}\).
\end{tablenotes}
\end{threeparttable}
\end{table*}

\begin{table*}[t]
\centering
\caption{The performance of sessions on CIFAR-100.}
\label{tab:cifar-sessions}
\begin{threeparttable}
\renewcommand{\arraystretch}{1.15}
\setlength{\tabcolsep}{5pt}
\begin{tabular}{lccccccccccc}
\toprule
\rowcolor{rowgray}
\textbf{Method} & 
\textbf{$S_{\mathrm{base}}$} & 
\textbf{1} & 
\textbf{2} & 
\textbf{3} & 
\textbf{4} & 
\textbf{5} & 
\textbf{6} & 
\textbf{7} & 
\textbf{$S_{\mathrm{last}}$} & 
\textbf{$S_{\mathrm{avg}}$} &
\textbf{PD} \\
\midrule

CEC\(\ddagger\) \cite{zhang2021few}
& 74.20 & 71.49 & 70.11 & 67.34 & 65.96 & 65.14 & 64.74 & 63.48 & 61.48 & 67.10 & 12.72 \\

L2P\(\ddagger\) \cite{wang2022learning}
& 83.29 & 76.81 & 71.29 & 66.53 & 62.38 & 58.68 & 55.42 & 52.49 & 49.87 & 64.08 & 33.42 \\

DualPrompt\(\ddagger\) \cite{wang2022dualprompt}
& 85.11 & 78.42 & 72.81 & 67.92 & 63.69 & 59.92 & 56.60 & 53.62 & 50.93 & 65.45 & 34.18 \\

NC-FSCIL\(\ddagger\) \cite{yang2023neural}
& 89.51 & 82.62 & 76.72 & 71.61 & 67.13 & 63.18 & 59.67 & 56.53 & 53.70 & 68.96 & 35.81 \\

WaRP\(\ddagger\) \cite{kim2023warping}
& 86.20 & 82.58 & 79.30 & 75.57 & 73.46 & 71.07 & 69.58 & 67.70 & 65.48 & 74.55 & 20.72 \\

PL-FSCIL\(\dagger\) \cite{yuan2025prompt}
& 85.73 & 74.54 & 74.77 & 72.42 & 72.98 & 72.87 & 72.49 & 71.62 & 75.13 & 74.73 & 10.60 \\

PriViLege\(\ddagger\) \cite{park2024pre}
& 90.88 & 89.39 & 88.97 & 87.55 & 87.83 & 87.35 & 87.53 & 87.15 & 86.06 & 88.08 & \textbf{4.82}\\

CKPD-FSCIL\(\ddagger\) \cite{li2025continuous}
& 91.57 & 90.03 & 89.84 & 88.44 & 88.58 & 87.74 & 87.82 & 87.36 & 86.22 & 88.62 & 5.35\\

\midrule

\rowcolor{rowgray}
\textbf{Proposed Method\(\dagger\)}
& 93.43 & 91.42 & 90.83 & 89.17 & 89.30 & 88.89 & 88.91 & 88.63 & 87.42 & 89.78 & 6.01 \\

\midrule

\rowcolor{rowgray}
\textbf{Proposed Method\(\ddagger\)}
& \textbf{93.92} & \textbf{92.12} & \textbf{91.67} & \textbf{90.48} & \textbf{90.60} & \textbf{90.21} & \textbf{90.16} & \textbf{89.83} & \textbf{88.75} & \textbf{90.86} & 5.17 \\

\bottomrule
\end{tabular}
\begin{tablenotes}
\footnotesize
\item \textbf{Note:} The columns represent the base session (\(S_{\mathrm{base}}\)), incremental sessions (1--7), the final session (\(S_{\mathrm{last}}\)), the average accuracy (\(S_{\mathrm{avg}}\)), and the performance drop (PD) between \(S_{\mathrm{base}}\) and \(S_{\mathrm{last}}\).
\end{tablenotes}
\end{threeparttable}
\end{table*}

\vspace{2mm}
\noindent \textbf{Incremental Sessions.} 
In the incremental sessions, new classes are introduced in a $N$-way $K$-shot manner, meaning that eash session presents a small set of new classes (way) that contain only a few labeled examples (shots). To mitigate overfitting and catastrophic forgetting, all parameters of the feature extractor remain frozen. Consequently, each few-shot sample is passed through the feature extractor to obtain its feature representation. For each new class $j'$, we define its weight vector $\mu_{j'}^P$ analogously to Eq. \ref{eq:prototype} using the available few-shot samples. We normalize $\mu_{j'}^P$ and append each new class to the classifier’s weight matrix:

\begin{equation}
\mu_{j'} \leftarrow \frac{\mu_{j'}^{P}}{\|\mu_{j'}^{P}\|_2}.
\end{equation}

\noindent By updating the classifier in this manner, we incorporate new classes with minimal disruption to the existing feature representations, thus maintaining robust performance on previously learned classes.

%In the incremental sessions, new classes are introduced in a $N$-way $K$-shot manner, meaning that eash session presents a small set of new classes (way) that contain only a few labeled examples (shots). To mitigate overfitting and catastrophic forgetting, all parameters of the feature extractor remain frozen. Consequently, each few-shot sample is passed through the feature extractor to obtain its feature representation. For each new class $j'$, its prototype $\phi_{j'}^P$ is computed analogously to Eq. 7 using the available few-shot samples. We normalize $\phi_{j'}^P$ and append each new class to the classifier’s weight matrix:

%\begin{equation}
%\phi_{j'} \leftarrow \frac{\phi_{j'}^{P}}{\|\phi_{j'}^{P}\|_2}.
%\end{equation}

%\noindent By updating the classifier in this manner, we incorporate new classes with minimal disruption to the existing feature representations, thus maintaining robust performance on previously learned classes.

%\ar{you don't explain how you train new parameters $\Delta W$. So everything is well explain so far, but you introduced these new parameters and never talked about them later. tell the cost function and how they are trained with few shots. I'm guessing here is the place to say it} 

\section{Experimental Results}
\label{sec:results}
\noindent \textbf{Datasets and Metrics.} 
Our method is evaluated on three widely used datasets in the FSCIL community: CUB-200 \cite{wah2011caltech}, CIFAR-100 \cite{krizhevsky2009learning}, and miniImageNet \cite{vinyals2016matching}. We follow the same splits as previously published manuscripts \cite{kim2023warping, yang2023neural, park2024pre, yuan2025prompt,zhang2021few, wang2022dualprompt, wang2022learning}. For CUB-200, the base session contains 100 classes, and each incremental session introduces 10 new classes (ways) with 5 examples per class (shots), resulting in one base session plus 10 incremental sessions. For CIFAR-100 and miniImageNet, the base session consists of 60 classes, while the incremental sessions adopt a 5-way 5-shot configuration, yielding one base session followed by 8 incremental sessions. The base session accuracy is reported as $S_{\mathrm{base}}$, the final session as $S_{\mathrm{last}}$, the average accuracy across all sessions as $S_{\mathrm{avg}}$, and the performance drop (PD) as the difference between $S_{\mathrm{base}}$ and $S_{\mathrm{last}}$ (lower the better).

\noindent \textbf{Implementation.}
In our model, we implement two ViT backbones. The first backbone is ViT-B/16 pre-trained on ImageNet-21k and fine-tuned on ImageNet-1K (denoted as \(\dagger\)). The second backbone is the more powerful ViT-B/16 pre-trained on ImageNet-21K (denoted as \(\ddagger\)). We focus solely on shared additive updates to the self-attention layers of all 12 transformer blocks, following \cite{hu2022lora}. For all datasets, images are uniformly resized to 224 × 224 pixels to meet the input requirements of the ViT architecture. Our method was trained for 10 epochs on both CUB-200 and CIFAR-100, and 3 epochs on miniImageNet. We used the SGD optimizer on a single NVIDIA L4 GPU and maintained a batch size of 128 across all datasets.

\subsection{Main Experimental Results}
\begin{figure}[t]
\centering
\includegraphics[width=\linewidth]{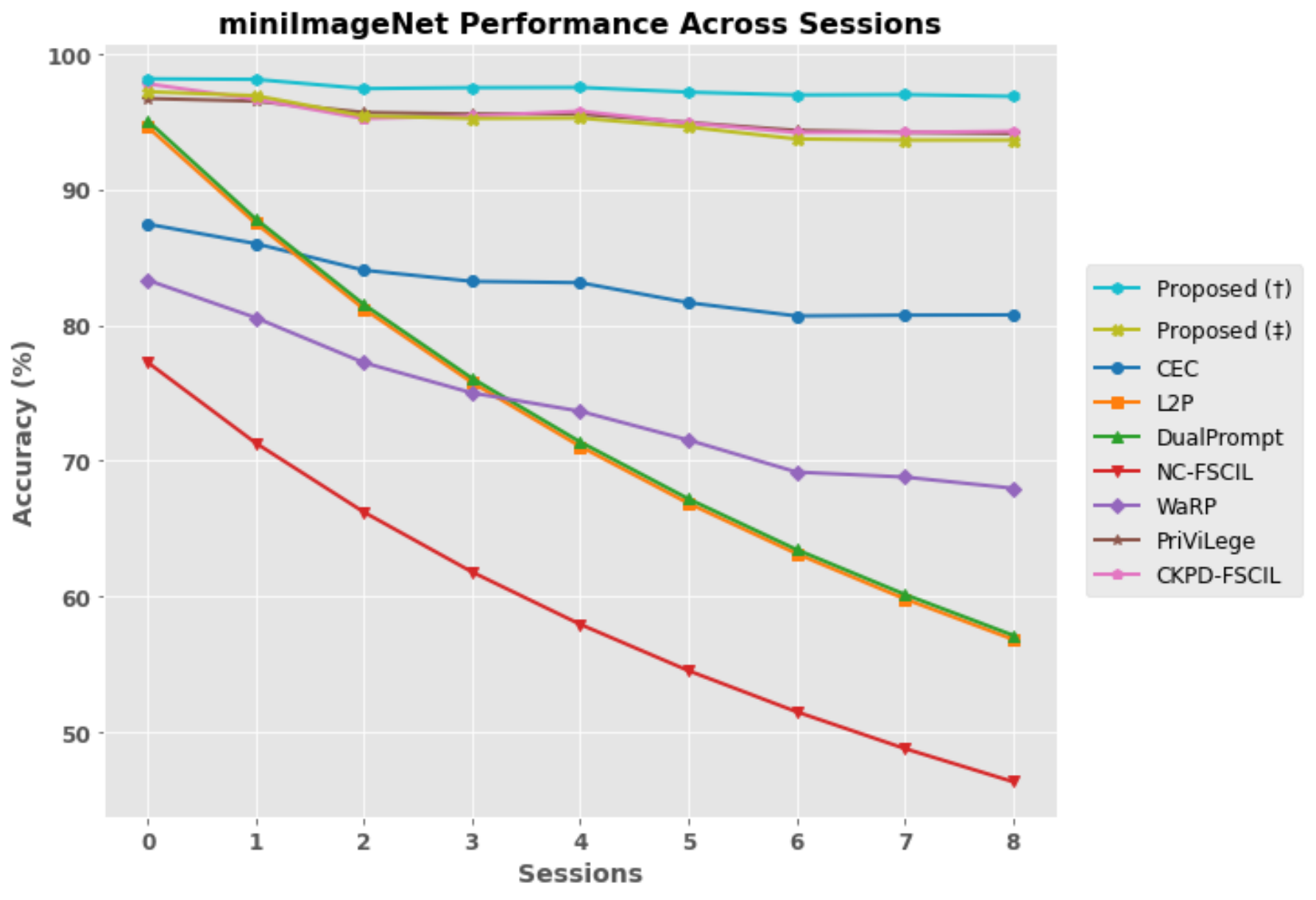}
\caption{miniImageNet performance across sessions.}
\label{fig:Mini}
\end{figure}

In this section, we discuss our main experimental results compared against baseline FSCIL methods. Session-wise results on CUB-200, CIFAR-100, and miniImagnet are shown in Tables \ref{tab:cub200-sessions}, \ref{tab:cifar-sessions}, and Figure \ref{fig:Mini}, respectively. 

On CUB-200, our method improves the base accuracy (\(S_{\mathrm{base}}\)) by 3.25\% and the final session accuracy (\(S_{\mathrm{last}}\)) by 9.18\% over the best competing methods, while reducing the performance drop (PD) by 0.66\%. These improvements show that our selective parameter updates not only preserve base knowledge but also better integrate new classes over incremental sessions. On CIFAR-100, we achieve state-of-the-art accuracies during both the base and all incremental sessions. Although our PD (5.17\%) is marginally higher than that of PriViLege (4.82\%), our method avoids the extra complexity and parameter overhead associated with prompt tuning and word embeddings required by PriViLege. For miniImageNet, our approach attains stellar performance across all sessions, setting a new benchmark PD of 1.29\%, a new base session accuracy of 98.13\% and final session accuracy of 96.84\%  when using the ViT backbone fine-tuned on ImageNet-1K. Interestingly, the less powerful backbone (with ImageNet-1K fine-tuning) outperforms its counterpart, likely due to better alignment with the miniImageNet distribution. Overall, these results validate that our FSCIL framework, with its targeted additive updates and refined classifier weights, effectively maintains robust representations while accommodating new classes, leading to significant improvements over existing methods.

\subsection{Ablation Study}

\noindent \textbf{Number of Updated Self-Attention Layers.} In the first ablation experiment, we study the effect of applying the additive update only to the final $n$ self-attention layers of each transformer block in the ViT architecture. Note that when the additive update is applied to 0 blocks, the model remains identical to the frozen, pre-trained ViT, and only the classifier is updated during training. 

\begin{table}[t]
\centering
\caption{Ablation Study: Impact of Applying Additive Updates to Only the Final $n$ Self-Attention Blocks}
\label{tab:ablation}
\begin{threeparttable}
\renewcommand{\arraystretch}{1.2}
\setlength{\tabcolsep}{6pt}
\begin{tabular}{lcc|cc}
\toprule
\multirow{2}{*}{\textbf{Blocks}} & \multicolumn{2}{c|}{\textbf{CUB-200}} & \multicolumn{2}{c}{\textbf{CIFAR-100}} \\
\cmidrule(lr){2-3}\cmidrule(lr){4-5}
 & $S_{\mathrm{avg}}$ (\%) & PD (\%) & $S_{\mathrm{avg}}$ (\%) & PD (\%) \\
\midrule
0  & 77.37 & 8.54 & 60.48 & 14.43 \\
4  & 79.11 & 6.63 & 77.65 & 13.66 \\
8  & 82.74 & 4.52  & 88.73 & 7.33 \\
12 & \textbf{84.97} & \textbf{4.15} & \textbf{90.86} & \textbf{5.17} \\
\bottomrule
\end{tabular}
\end{threeparttable}
\end{table}

Table~\ref{tab:ablation} summarizes our findings on the CUB-200 and CIFAR-100 datasets. As we gradually increase the number of self-attention layers receiving additive updates, the model exhibits improved performance. On CUB-200, the average accuracy (\(S_{\mathrm{avg}}\)) increases from 77.37\% (with 0 updated blocks) to 84.97\% (with all 12 blocks updated), resulting in a gain of 7.60 percentage points, while the PD is reduced by over 4 percentage points. Similarly, on CIFAR-100, \(S_{\mathrm{avg}}\) improves significantly from 60.48\% to 90.86\%, and PD decreases from 14.43\% to 5.17\%. The more pronounced improvements on CIFAR-100 can be attributed to its dataset split and incremental session configuration, which features fewer training base classes and a 5-way 5-shot format in each session, thereby making it more susceptible to catastrophic forgetting. These results suggest that updating a greater number of self-attention layers significantly enhances the model's adaptability while preserving the robust representations learned during the base session.

\vspace{1mm}
\noindent \textbf{Impact of Additive Updates on MLP Layers.} In this experiment, we evaluate the impact of applying additive updates to the feed-forward (MLP) layers within each transformer block. Converse to our baseline, we instead freeze the self-attention layers and apply the additive update only on the MLP layers, then compare the resulting performance differences. Table \ref{tab:comparison_update} presents a comparison between self-attention and MLP update strategies on CUB-200 and CIFAR-100 datasets. 

\begin{table}[t!]
\centering
\caption{Comparison of Self-Attention and MLP Update Strategies on CUB-200 and CIFAR-100}
\label{tab:comparison_update}
\renewcommand{\arraystretch}{1.1}
\setlength{\tabcolsep}{8pt}
\begin{tabular}{lcc|cc}
\toprule
Metric & \multicolumn{2}{c|}{CUB-200} & \multicolumn{2}{c}{CIFAR-100} \\
\cmidrule(lr){2-3}\cmidrule(lr){4-5}
       & Self-Attn & MLP & Self-Attn & MLP \\
\midrule
\(S_{\mathrm{base}}\) & 88.41 (\textcolor{blue}{+0.56}) & 87.85 & 93.92 (\textcolor{red}{-0.55}) & 94.37 \\
\(S_{\mathrm{last}}\) & 84.26 (\textcolor{blue}{+0.85}) & 83.41 & 88.75 (\textcolor{blue}{+0.56}) & 88.19 \\
\(S_{\mathrm{avg}}\)  & 84.97 (\textcolor{blue}{+0.80}) & 84.17 & 90.86 (\textcolor{blue}{+0.01}) & 90.85 \\
PD                   & 4.15 (\textcolor{blue}{+0.29}) & 4.44  & 5.17 (\textcolor{blue}{+1.01}) & 6.18 \\
\bottomrule
\end{tabular}
\end{table}

On CUB-200, the self-attention update achieves a base accuracy of 88.41\% (0.56 points higher), a final session accuracy of 84.26\% (0.85 points higher), and a PD of 4.15\% (0.29 points lower) compared to the MLP update. On CIFAR-100, although the MLP update shows a slightly higher base accuracy (94.37\% vs. 93.92\%), the self-attention update attains a higher final accuracy (88.75\% vs. 88.19\%) and a notably lower PD (5.17\% vs. 6.18\%). Notably, the self-attention update strategy also introduces fewer trainable parameters, $3d^2$, since it only modifies the QKV projection matrices. However, updating both fully connected layers in the MLP module introduces approximately $8d^2$ parameters, significantly increasing the parameter count. %This reduced parameter overhead makes the self-attention approach more efficient and less prone to overfitting in few-shot incremental learning settings.

\section{Conclusion}
\label{sec:conclusion}
In this paper, we proposed a novel FSCIL framework that leverages a frozen pre-trained ViT backbone augmented with parameter-efficient additive updates. By selectively updating only the self-attention layers while keeping the pre-trained weights frozen, our method provides a simple strategy for effectively integrating new classes with only few-shot samples while preserving robust base representations. Extensive experiments on CUB-200, CIFAR-100, and miniImageNet demonstrate that our approach significantly outperforms previous state-of-the-art methods in terms of base accuracy, incremental accuracy, and performance drop. Our findings underscore the advantages of targeted parameter tuning in mitigating catastrophic forgetting and reducing overfitting in FSCIL. %In future work, we plan to explore additional PEFT techniques to further advance this challenging problem.

\ifCLASSOPTIONcaptionsoff
  \newpage
\fi

\bibliographystyle{IEEEtran}% or another style like plain, alpha, etc.
\bibliography{ref} 

\end{document}